\documentclass[runningheads]{llncs}
\usepackage{graphicx}
\usepackage{caption}
\usepackage{multirow}
\usepackage{svg}
\usepackage{amsmath}
\usepackage{subcaption}
\usepackage{pgfplots}
\usepackage{circuitikz}
\usepackage[altpo]{backnaur}
\usepackage{url}
\usepackage{amssymb}

\usepackage{pgfplotstable}
\pgfplotstableread{
	Label s1 s2 s3 s4 s5
	STEM 1 4 4 7 7
	S-ENN 0 0 3 0 7
	Mixup 1 4 0 1 1
	ADA 1 4 0 1 0
	S-NC 1 4 0 1 0
	SVM-S 1 0 1 1 2
	S-Tomek 1 4 0 1 0
	BSMOTE 1 0 0 1 0
	SMOTE 1 0 0 1 0
}\testdata

\pgfplotsset{width=12cm,height=6cm}

\usepackage{tikz}

\usepackage{xcolor}
\definecolor{grey}{HTML}{D4D4D4}
\definecolor{quincy}{HTML}{ef476f}
\definecolor{banana}{HTML}{ffd166}
\definecolor{neptune}{HTML}{06d6a0}
\definecolor{tango}{HTML}{118ab2}
\definecolor{fuel}{HTML}{8c8c8c}

\definecolor{c1}{RGB}{0, 0, 0}
\definecolor{c2}{RGB}{230, 159, 0}
\definecolor{c3}{RGB}{86, 180, 233}
\definecolor{c4}{RGB}{0, 158, 115}
\definecolor{c5}{RGB}{0, 114, 178}
\definecolor{c6}{RGB}{213, 94, 0}
\definecolor{c7}{RGB}{204, 121, 167}
\definecolor{c8}{RGB}{240, 228, 65}
\pgfplotsset{compat=1.18} 

\begin{document}
\title{Interpretable Solutions for Breast Cancer Diagnosis with Grammatical Evolution and Data Augmentation}
%
%
\author{Yumnah Hasan\inst{1}\orcidID{0000-0001-9310-8886} \and Allan de Lima\inst{1}\orcidID{0000-0002-1040-1321} \and Fatemeh Amerehi\inst{1}\orcidID{0000-0002-6255-4573} \and \\Darian  Reyes Fern\'{a}ndez de Bulnes\inst{1}\orcidID{0000-0002-7413-5122} \and \\Patrick Healy\inst{1}\orcidID{0000-0002-3824-7442}\and  Conor Ryan \inst{1}\orcidID{0000-0002-7002-5815}}
\authorrunning{Y. Hasan et al.}
%
\institute{\textsuperscript{1}University of Limerick, Limerick, Ireland
\email{\{Yumnah.Hasan,Allan.Delima,Fatemeh.Amerehi,Darian.Reyesfernandezdebulnes,\\Patrick.Healy,Conor.Ryan\}@ul.ie}}
\maketitle              

\begin{abstract}
Medical imaging diagnosis increasingly relies on Machine Learning (ML) models. This is a task that is often hampered by severely imbalanced datasets, where positive cases can be quite rare. Their use is further compromised by their limited interpretability, which is becoming increasingly important. While \textit{post-hoc} interpretability techniques such as SHAP and LIME have been used with some success on so-called black box models, the use of inherently understandable models makes such endeavours more fruitful. This paper addresses these issues by demonstrating how a relatively new synthetic data generation technique, STEM, can be used to produce data to train models produced by Grammatical Evolution (GE) that are inherently understandable. STEM is a recently introduced combination of the Synthetic Minority Oversampling Technique (SMOTE), Edited Nearest Neighbour (ENN), and Mixup; it has previously been successfully used to tackle both between-class and within-class imbalance issues. We test our technique on the Digital Database for Screening Mammography (DDSM) and the Wisconsin Breast Cancer (WBC) datasets and compare Area Under the Curve (AUC) results with an ensemble of the top three performing classifiers from a set of eight standard ML classifiers with varying degrees of interpretability. We demonstrate that the GE-derived models present the best AUC while still maintaining interpretable solutions.

\end{abstract}

\section{Introduction}

In medical imaging diagnoses, where decisions can have significant implications for individual's health, it is essential to gain a thorough understanding of the factors influencing these decisions. While Machine Learning (ML) models have proven effective in diagnosing a variety of medical conditions in medical imaging~\cite{Varoquaux}, their limited interpretability poses a challenge to their broader adoption. Moreover, the recently introduced European Union (EU) Communication on Fostering a European approach toAI~\cite{CommunicationFosteringEuropean2021} specifically targets explainability as a key concern for the deployment of ML and Artificial Intelligence (AI) models. 

Another prevalent challenge in the medical imaging domain is the issue of class imbalance within the dataset. Methods such as Synthetic Minority Oversampling Technique (SMOTE), Edited Nearest Neighbour (ENN), and Mixup combined together as STEM~\cite{STEM}, which leverages the full distribution of minority classes, can effectively address both inter-class and intra-class imbalances. In~\cite{STEM}, STEM was applied in-conjunction with an ensemble of ML classifiers, producing promising outcomes. However, understanding the reasoning behind ML model predictions remains a complex task. Furthermore, as the volume of instances and the specificity of problems grow, the complexity of the derived solutions also increases.

Building trust in ML classifiers and understanding the behaviour of the solutions is pivotal to their broader acceptance. Employing inherently explainable models is a useful strategy when generating Explainable AI models. Grammatical Evolution (GE)~\cite{Ryan}, an Evolutionary Computation (EC) technique, has been used to leverage grammars to define and constrain the syntax of potential solutions, producing inherently explainable models~\cite{murphy2021towards}.

To address these challenges, we developed a classification system based on GE. Our study includes a comprehensive comparison with an ensemble of other ML classifiers. Notably, GE models show enhanced interpretability compared to other traditional ML models. GE provide solutions in the form of symbolic expressions, offering a more intuitive understanding of the decision-making process. This emphasis on interpretability is crucial, especially in healthcare, where understanding the rationale behind decisions is of paramount importance.

Our research hypothesises that the use of the STEM augmentation technique combined with an approach rooted in GE produces more interpretable solutions as compared to the other ensemble ML classifiers. 

The contributions of this paper are as follows. Firstly, we develop a method that combines a GE classifier with STEM, outperforming an ensemble of ML classifiers, as indicated by the superior AUC. Secondly, our approach distinguishes itself by offering more interpretable solutions compared to the ensemble method.
Finally, the paper presents rigorous statistical analyses to comprehensively evaluate the performance of implemented data augmentation techniques on each data setup. 

The rest of the paper is structured as follows:
Section~\ref{Literature} reviews the existing literature. Section~\ref{Methodology} outlines the proposed methodology, and Section~\ref{Experiments} addresses experimental details performed in this work. Results and discussion are described in Section~\ref{Results}, and Section~\ref{Conclusion} presents the conclusion and future guidelines.


\section{Literature Review}
\label{Literature}

In the realm of medical applications, particularly in the context of breast cancer diagnosis, the issue of imbalanced datasets is a critical concern. Imbalances, where one class significantly outweighs the other, can introduce biases and compromise the reliability of ML models. Implementing effective strategies for class balancing, such as oversampling, undersampling, and their combination, results in a more balanced and representative training dataset~\cite{fernandez2013analysing}. Previous studies~\cite{he2008adasyn,han2005borderline} have recognized the impact of class imbalance in medical datasets for ML tasks.

Moreover, ML algorithms have demonstrated notable efficiency in the classification of medical data. A compelling study showcases the effectiveness of ensembles, where Bayesian networks and Radial Basis Function (RBF) classifiers with majority voting resulted in an accuracy of 97\% \cite{Jabbar} when applied to the Wisconsin Breast Cancer (WBC) dataset. Furthermore, an approach that combined linear and non-linear classifiers using Micro Ribonucleic Acid (miRNA) profiling achieved an impressive accuracy of 98.5\% \cite{sharma}. 

While these findings are promising, ML algorithms may struggle to contextualize information and are susceptible to unexpected or undetected biases originating from input data. Additionally, they often lack transparent justifications for their predictions or decisions \cite{B.M}. In response to this, employing GE can yield interpretable solutions. As a variant of Genetic Programming, GE evolves human-readable solutions, offering explanations for the rationale behind its classification decisions, which is a significant advantage over current paradigms in unsupervised and semi-supervised learning \cite{Fitzgerald}.

Previous studies have already demonstrated the effectiveness of GE across a range of ML tasks. It has proven valuable for feature generation and feature selection \cite{Gavrilis}, as well as for hyperparameter optimization \cite{Noorian}. The GenClass system~\cite{Anastasopoulos}, built upon GE, demonstrates promising outcomes and outperforms RBF networks in certain classification problems. They utilized thirty benchmark datasets from the UCI and KEEL repositories, including Haberman, which consists of breast cancer instances. While it has excelled in these areas, there are still avenues for further exploration.

In this paper, we aim to investigate the efficiency of utilizing GE as a medical imaging classifier combined with STEM to handle imbalance distributions of data samples, particularly in breast cancer diagnosis. Leveraging the interpretive and adaptable features of GE, our objective is to achieve accurate and reliable outcomes that can be easily explained.


\section{Methodology}
\label{Methodology}

For analysis, we utilize two primary breast cancer datasets. One consists of images, the Digital Database for Screening Mammography (DDSM)~\cite{heath1998current}
, while the other consists of tabular data, the WBC~\cite{wolberg1992breast} dataset. $DDSM$ is a comprehensive collection of mammograms, encompassing both normal and abnormal images. For this study, we focused on $DDSM's$ Cancer 02 volume and three volumes of normal samples (Volume 01-03). By selecting
one volume of cancer images compared to three volumes of normal images, we maintain a realistic class imbalance ratio. These images come from the Craniocaudal (CC) and Mediolateral Oblique (MLO) views of both the left and right breasts. We work with 152 cancerous images and 876 healthy ones from volumes 1-3. Each image was divided into four segments: the entire breast (\textbf{I}), the top segment (\textbf{It}), the middle segment (\textbf{Im}), and the bottom segment (\textbf{Ib}).

\begin{figure}[!ht]
    \centering
    \includegraphics[width=\columnwidth]{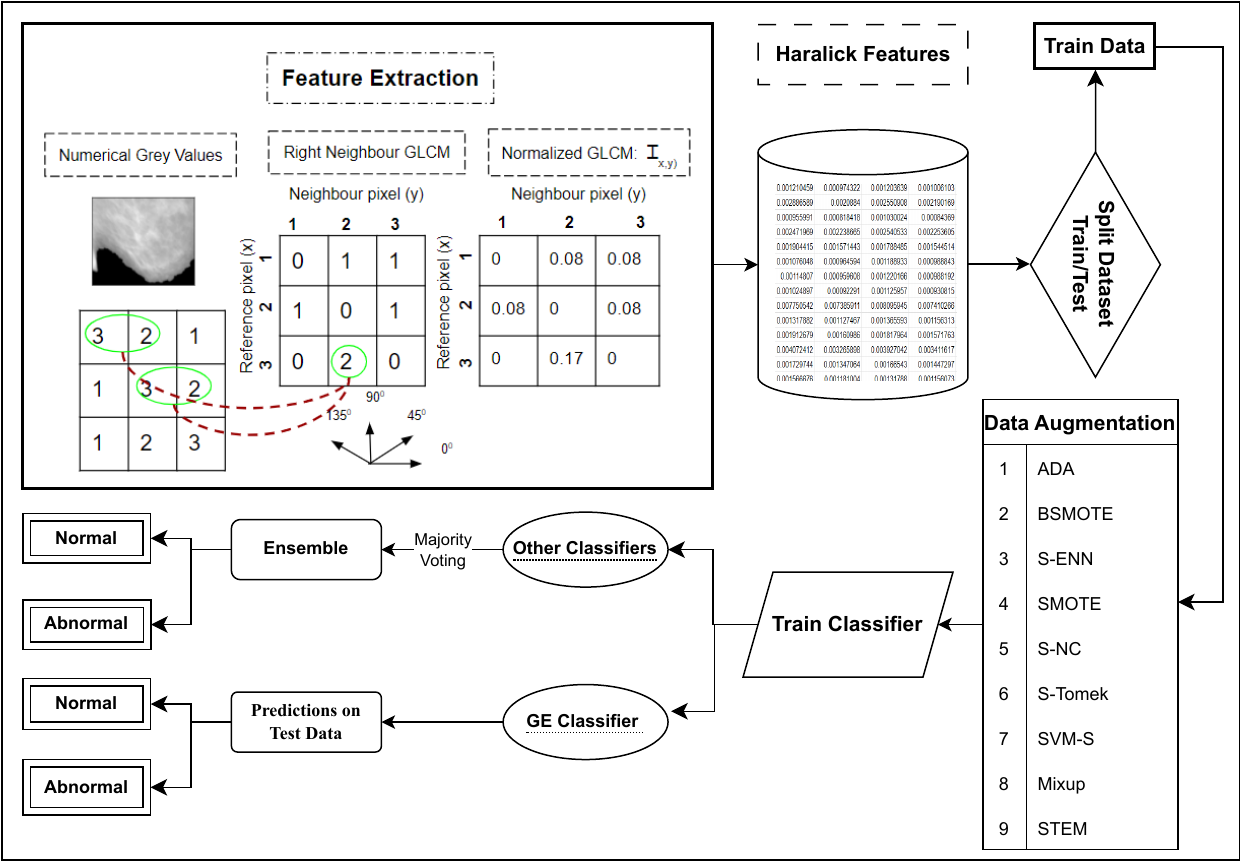}
     \captionsetup{justification=centering}
    \caption{Outline of the proposed approach for breast cancer classification using GE and other classifiers.}
    \label{methodology}
\end{figure}

The \textit{WBC} dataset consists of 30 features derived from Fine Needle Aspiration (FNA) samples of breast masses, categorising patients into benign (non-cancerous) and malignant (cancerous) cases. It comprises 212 malignant samples and 357 benign samples.

To create a dataset containing breast cancer images from the $DDSM$ image for evaluating the proposed methodology, we first need to extract features that will be used for training. This involves isolating the breast region, eliminating irrelevant background data, segmenting the breast region, and extracting pertinent features to generate a comprehensive training dataset of breast segments. Initially, a median filter is applied to reduce noise within the images. Subsequently, non-essential background data, often containing machine-generated labels such as `CC' or `MLO', is removed. For this step, we employed a precise Otsu thresholding technique. Following this, the segmenting process proposed in~\cite{RyanB1} effectively partitioned images into three overlapping segments.

Feature extraction is the next critical phase. In our study, we extracted a set of Haralick's Texture Features~\cite{haralick1973textural} from both whole and segmented images. The selection of these features is based on the hypothesis that there are discernible textural differences between normal and abnormal images. Specifically, we compute thirteen distinct Haralick features from the Gray-Level Co-Occurrence (GLCM) matrix, employing four orientations corresponding to two diagonal (grey-level numeric values of the images) and two adjacent neighbours. This process results in generating a total of 52 features per segment or image.

High class imbalance present in the utilized datasets poses a significant challenge in developing robust and accurate predictive models.  Therefore, explicit data augmentation has been implemented in the training set to effectively address this class imbalance challenge. Using nine distinct augmentation approaches outlined in Section~\ref{Oversampling}, synthetic samples are generated to enrich the dataset with more discriminative information, ultimately improving the learning capabilities of the model. 

In the last step, the GE classifier and an ensemble of other ML classifiers are trained separately to make predictions on the test set. Augmented training data is used, while the original imbalanced test set is used for testing. For ensembling, eight ML classifiers are used as mentioned in Section~\ref{classifiers}. The top three classifiers, based on AUC, are selected and combined through majority voting to create the final predictions. The complete pipeline of the proposed approach is shown in Fig.~\ref{methodology}.


\section{Experimental Details}
\label{Experiments} 

The DDSM and WBC datasets are used to evaluate the proposed technique. The study employs five different data setups to train the classifiers. For the WBC dataset, a single setup is utilized, consisting of 30 breast mass features per sample acquired through FNA.

In contrast, the DDSM dataset includes images from two views, CC and MLO. To conduct experiments, the dataset is categorized into four distinct configurations based on these views. In the initial setup, denoted as  ``$S_{CC}$'', data is exclusively extracted from segments of the CC view. Conversely, the second category,  ``$S_{MLO}$'', comprises segmented images exclusively from the MLO view. The third configuration,  ``$S_{CC+MLO}$'', combines segments from both views. Lastly, the fourth setup,  ``$F_{CC+MLO}$'', considers the full image (non-segmented) features from both the CC and MLO views for comprehensive analysis. The number of features for each segment or image is 52, used in all these setups

\begin{figure}[!ht]
\centering
\begin{tikzpicture}
\foreach [count=\i] \total/\clr in {7/grey,6/grey,6/grey,6/grey,37/grey}
   \draw [line width=2mm,\clr]
     (1.5cm+\i*3.5mm,0)
     node[above,inner sep=0pt,black,font=\scriptsize]{\total}
     arc[start angle=0,radius=0.8cm+\i*3.5mm,delta angle=-360];
\foreach [count=\i] \total/\clr in {7/fuel,6/banana,6/neptune,6/quincy,37/tango}
   \draw [line width=2mm,\clr]
     (1.5cm+\i*3.5mm,0)
     node[above,inner sep=0pt,black,font=\scriptsize]{\total}
     arc[start angle=0,radius=0.8cm+\i*3.5mm,delta angle=-3.60*\total];
\matrix [every node/.style={right=1.5mm,black,font=\normalsize}] at (6.5,1) {
   \fill[fuel] circle[radius=2mm] node  {$DDSM - S_{CC}$ (85:1316)}; \\
   \fill[banana] circle[radius=2mm] node  {$DDSM - S_{MLO}$ (89:1326)}; \\
   \fill[neptune] circle[radius=2mm] node  {$DDSM - S_{CC + MLO}$ (174:2642)}; \\
   \fill[quincy] circle[radius=2mm] node  {$DDSM - F_{CC + MLO}$ (152:876)}; \\
   \fill[tango] circle[radius=2mm] node  {\textit{WBC} (212:357)}; \\
};
\end{tikzpicture}
\caption{Concentric ring chart for setup description. Rings are setups, and the coloured areas indicate training positive percent. Legend includes the training positive and negative total samples.} 
\label{fig2}
\end{figure}
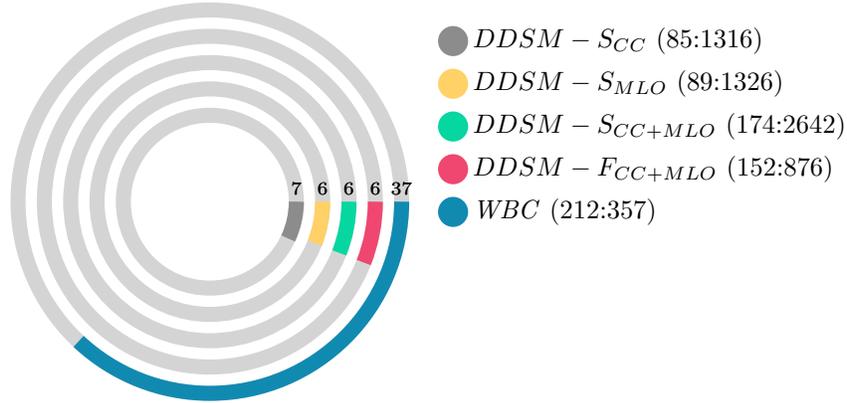

We divided the datasets into training and testing sets at an 80:20 ratio, respectively. Notably, all $DDSM$ configurations exhibit significant class imbalances, with class ratios ranging from 6:94 $S_{CC}$ , $S_{MLO}$ and $S_{CC+MLO}$ setups. For $F_{CC+MLO}$ the ratio between the positive versus negative class is 15:85. Likewise, the \textit{WBC} dataset has a class distribution of 37\% positive and 63\% negative classes as illustrated in Fig.~\ref{fig2}. 

\subsection{System Settings}
All the ML experiments were conducted using the PyCaret library~\cite{PyCaret}. The GRAPE~\cite{Allan} framework was used to perform GE experiments. For statistical analysis, we employed the AutoRank Python library~\cite{Herbold} to evaluate the performance of the implemented augmentation approaches. Our code, as well as our dataset configurations, are available in our supplementary material.

\subsection{Performance Metric}
To evaluate the performance of the designed approach, AUC has been selected as the assessment metric which uses Trapezoidal rule for its computation. AUC has become a widely accepted performance measure in classification problems due to its reliability, particularly in the context of imbalanced datasets~\cite{liang2020lr,Halimu}.AUC serves as a comprehensive metric, encompassing both sensitivity (Equation~\ref{eq2}) and specificity (Equation~\ref{eq3}), considering various threshold values. T\textsubscript{Pos} denotes true positives, T\textsubscript{Neg} true negatives, F\textsubscript{Pos} false positives, and F\textsubscript{Neg} denotes false negatives.

\begin{equation}
\label{eq2}
Sensitivity = \frac{T_{Pos}}{T_{Pos} + F_{Neg}}
\end{equation}
\begin{equation}
\label{eq3}
Specificity = \frac{T_{Neg}}{T_{Neg} + F_{Pos}}
\end{equation}

\subsection{Class Balancing}
\label{Oversampling}

The methods utilized for generating synthetic data with the aim of equalizing the class distribution ratio include the Synthetic Minority Oversampling Technique (SMOTE)~\cite{chawla2002smote}, Borderline SMOTE (BSMOTE)~\cite{han2005borderline}, SMOTENC (S-NC)~\cite{chawla2002smote}, Support Vector Machine SMOTE (SVM-S)~\cite{nguyen2011borderline}, Mixup~\cite{zhang2017mixup}, and ADASYN (ADA)~\cite{he2008adasyn}. Additionally, three hybrid methods, SMOTE Edited Nearest Neighbour (S-ENN)~\cite{wilson1972asymptotic} SMOTE-Tomek (S-Tomek)~\cite{batista2003balancing} and combination of SMOTE, ENN, and Mixup (STEM) are also implemented to compare against each other. 

Notably, STEM generates a balanced number of samples for each class. Compared to other methods, it demonstrates the ability to increase the number of data samples more extensively, resulting in improved model performance.

\subsection{Grammatical Evolution}

GE's grammars are typically defined in Backus-Naur Form (BNF), a notation represented by the tuple $N$, $T$, $P$, $S$, where $N$ is the set of $non-terminals$, transitional structures usually with semantic meaning, $T$ is the set of $terminals$, items in the phenotype, $P$ is a set of production rules, and $S$ is a start $non-terminal$. The following simple grammar was created to evolve solutions for the first four data setups with 52 numerical features, whereas, for the last setup, 30 numerical features were used:

\begin{bnf*}
  \bnfprod{expression}
    {\bnfpn{operator} \bnfts{(} \bnfpn{expression} \bnfts{,} \bnfpn{expression} \bnfts{} \bnfor \bnfpn{operand}}\\
  \bnfprod{operator}
        {\bnfts{add} \bnfor \bnfts{sub} \bnfor \bnfts{mul} \bnfor \bnfts{pdiv}}\\
  \bnfprod{operand}
        {\bnfpn{x} \bnfor \bnfpn{digit} \bnfpn{digit} \bnfts{.} \bnfpn{digit} \bnfpn{digit}}\\
  \bnfprod{x}
        {\bnfts{x[0]} \bnfsk \bnfts{x[51]}}\\
  \bnfprod{digit}
        {\bnfts{0} \bnfor \bnfts{1} \bnfor \bnfts{2} \bnfor \bnfts{3} \bnfor \bnfts{4} \bnfor \bnfts{5} \bnfor \bnfts{6} \bnfor \bnfts{7} \bnfor \bnfts{8} \bnfor \bnfts{9}}
\end{bnf*}

This grammar permits the use of basic arithmetic operations (addition, subtraction, multiplication, and division --protected in case the divisor is equal to $0$)  and the inclusion of real numbers constants. These constants are helpful because GE can explore beyond the parameter space given to minimize the error between expected and predicted outputs, something that does not happen with other ML classifiers. The $non-terminal$ $X$ encompasses the fifty-two numerical features for the first four setups of the DDSM dataset and the thirty numerical features for the WBC dataset.

The output domain of the evaluations is $o \in [-\infty,\infty]$. Subsequently, a sigmoid function is applied to constrain the values to $\sigma(o) \in [0, 1]$. For binary classification, the typical interpretation of the sigmoid function is the probability of belonging to class 1, and therefore we use $\sigma(o)$ to calculate AUC. Table~\ref{GE experiments} presents the experimental parameters used in this work:

\begin{table}
    \centering
    \caption{List of parameters used to run GE}
    \label{GE experiments}
    \begin{tabular}{lc}
    \hline Parameter type & Parameter value \\
    \hline Number of runs & 30 \\
    Number of generations & 100 \\
    Population size & 200 \\
    Mutation probability & 0.01 \\
    Crossover probability & 0.8 \\
    Elitism size & 1 \\
    Codon size & 255 \\
    Initialisation & Sensible \\
    Maximum initial depth & 10 \\
    Maximum depth & 35 \\
    Wrapping & 0 \\
    \hline
    \end{tabular}%
\end{table}

\subsection{Other Classifiers}
\label{classifiers}

We also used the augmented training data to train a diverse ensemble of eight ML classifiers. This ensemble includes Random Forest (RF), Linear Discriminant Analysis (LDA), Quadratic Discriminant Analysis, LightGBM, XGBoost, AdaBoost, KNN, and Extra Trees models. Initially, a comprehensive model is trained using all eight classifiers. Subsequently, based on the AUC metric, the three best-performing models are selected. These selected models are then combined through a majority voting approach. The final predictions are made on the test dataset, which consists of imbalanced and unseen samples. 


\section{Results and Discussion}
\label{Results}

To evaluate the performance of the proposed method, five distinct data setups are employed. Four configurations are derived from the DDSM dataset, considering variations in views, segments, and full images. The fifth setup is from the WBC dataset. To enhance the robustness of the training setups, nine augmentation approaches are applied and compared. The assessment is conducted using an ensemble of other ML classifiers, alongside GE.

The performance of the classifiers is compared based on AUC for each dataset. The ensemble classifiers are denoted by their respective initials: $L_d$ for Linear Discriminant Analysis, $Q$ for Quadratic Discriminant Analysis, $E$ for ExtraTree, $R$ for Random Forest, $L_i$ for Lightgbm, $K$ for KNN, $A$ for Adaboost, and $X$ for Xgboost. It is important to note that the AUC values of the other ensemble classifiers are presented for a single run, and they are then compared against the median AUC derived from 30 runs conducted with GE.

Table~\ref{AUC} provides an overview of the results. In the first setup, $S_{CC}$, an AUC of 0.91 was achieved, outperforming the ensemble of $L_dQE$, which obtained an AUC of 0.90. Similarly, in the second setup, $S_{MLO}$, an AUC of 0.90 was attained, while the ensemble of $L_dQE$ achieved a slightly lower AUC of 0.84.

For the third setup $S_{CC+MLO}$, an AUC of 0.92 was observed using the GE classifier, outperforming other classifiers that yielded the highest AUC of 0.87 using $L_dQE$. When the classifiers were trained on full image features in setup $F_{CC+MLO}$, the highest AUC values were 0.94 and 0.85, obtained by the GE classifier and the ensemble of $L_dQE$, respectively.

When comparing the AUC using the \textit{WBC} dataset, both GE and the ensemble of $AKL_r$ achieved an AUC of 0.99. 

\begin{figure}[!ht]
    \centering
    \includegraphics[width=\columnwidth]{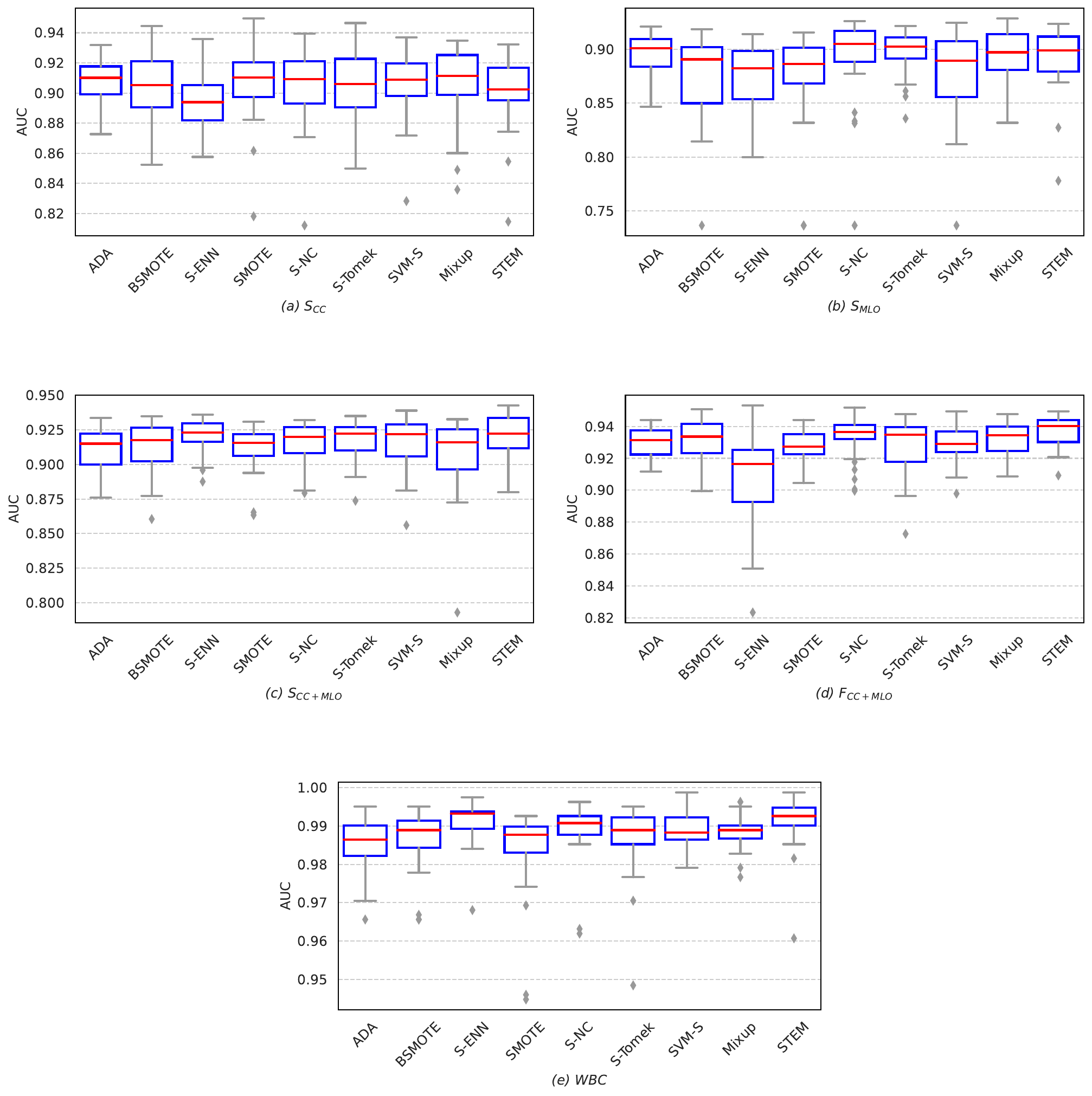}
     \captionsetup{justification=centering}
    \caption{Boxplot analysis comparing opponent approaches and their AUC distributions across multiple runs}
    \label{box_plot}
\end{figure}

\begin{table}[]
\caption{A comparison of the AUC for GE and the ensemble approaches using the nine different augmentation techniques for each data setup.}
\label{AUC}
\resizebox{\columnwidth}{!}{%
\begin{tabular}{ccccccccccc}
\hline
\multirow{2}{*}{Setups}       & \multirow{2}{*}{Classifiers} & \multirow{2}{*}{ADA} & \multirow{2}{*}{BSMOTE} & \multirow{2}{*}{S-ENN} & \multirow{2}{*}{SMOTE} & \multirow{2}{*}{S-NC} & \multirow{2}{*}{S-Tomek} & \multirow{2}{*}{SVM-S} & \multirow{2}{*}{Mixup} & \multirow{2}{*}{STEM} \\
                              &                              &                      &                         &                        &                        &                       &                          &                        &                        &                       \\ \hline
\multirow{3}{*}{$S_{CC}$}     & GE                          & 0.91                     & 0.90                        & 0.89                       & 0.91                       & 0.90                      & 0.90                          & 0.90                       & 0.91                       &  0.90                     \\
                              & \multirow{2}{*}{Others}        & 0.76                 & 0.73                    & 0.93                   & 0.77                   & 0.82                  & 0.77                     & 0.73                   & 0.90                   & 0.90                  \\
                              &                              & $L_dQE$                 & $L_dQE$                   & $L_dQE$                   & $L_dQE$                   & $L_dQE$                  & $L_dQE$                     & $L_dQE$                   & $L_dQE$                   & $L_dQE$                  \\ \hline
\multirow{3}{*}{$S_{MLO}$}    & GE                          & 0.90                     &    0.90                     & 0.90                       &    0.90                    & 0.87                      &     0.90                     & 0.89                       &   0.90                     &  0.89                     \\
                              & \multirow{2}{*}{Others}        & 0.80                 & 0.80                    & 0.80                   & 0.82                   & 0.78                  & 0.82                     & 0.81                   & 0.81                   & 0.84                  \\
                              &                              & $EL_iR$                 & $EL_iR$                    & $L_dQE$                   & $EL_iR$                   & $EL_iX$                 & $EL_iX$                     & $EL_iR$                   & $L_dQE$                   & $L_dQE$                  \\ \hline
\multirow{3}{*}{$S_{CC+MLO}$} & GE                          &  0.91                    &    0.91                     & 0.92                       &     0.91                   &  0.92                     &      0.91                    & 0.91                       &   0.90                     &  0.91                     \\
                              & \multirow{2}{*}{Others}        & 0.75                 & 0.68                    & 0.77                   & 0.75                   & 0.70                  & 0.76                     & 0.62                   & 0.76                   & 0.87                  \\
                              &                              & $EL_iR$                & $EL_iR$                    & $EL_iR$                   &$EL_iR$                   & $EL_iX$                 & $EL_iR$                     & $EL_iR$                   & $EL_iR$                   & $L_dQE$                 \\ \hline
\multirow{3}{*}{$F_{CC+MLO}$} & GE                          &    0.93                  & 0.91                       &   0.90                     &    0.92                    &  0.93                     &   0.94                       &  0.93                      &   0.93                     & 0.93                      \\
                              & \multirow{2}{*}{Others}        & 0.78                 & 0.84                    & 0.72                   & 0.81                   & 0.82                  & 0.82                     & 0.82                   & 0.81                   & 0.85                  \\
                              &                              & $EQR$                  & $EL_iR$                    & $ERX$                    & $EQR$                    & $EL_iQ$                  & $EL_iR$                     & $EQR$                    &  $L_iQL_d$                  & $L_dQE$                  \\ \hline
\multirow{3}{*}{\textit{WBC}}          & GE                          &  0.98                    & 0.98                        & 0.99                       & 0.98                       & 0.99                      &  0.98                        & 0.99                       & 0.98                       & 0.99                      \\
                              & \multirow{2}{*}{Others}        & 0.94                 & 0.94                    & 0.94                   & 0.94                   & 0.95                  & 0.94                     & 0.94                   & 0.94                   & 0.99                  \\
                              &                              & $L_dQE$                 & $L_dQE$                   &  $EKL_i$                   & $L_dQE$                   & $L_dQE$                  & $L_dQE$                    & $L_dQE$                   & $L_dEL_i$                  & $AKL_r$                  \\ \hline
\end{tabular}%
}
\end{table}
The augmentation approaches
are compared using the boxplot presented in Fig.~\ref{box_plot}. The plot indicates the AUC obtained from all nine augmentation approaches for each setup across all 30 runs. The horizontal line in red indicates the median value of the respective group. 

GE provides valuable insights into the most informative features used in the solutions, as demonstrated in Table~\ref{feature_Analysis}, which present the most frequently used features for each setup. The features extracted and presented in these tables are sorted by their impact on the solutions. Common features consistently found in Table \ref{feature_Analysis} for the $DDSM$ dataset include ``Inverse Difference Moment (IDM)''(feature 17), ``Contrast'' (feature 5),  and ``Difference Entropy'' (feature 41). Both contrast and IDM represent the difference in grey levels between pixels, while entropy indicates the level of randomness in the grey levels.

For the \textit{WBC} dataset, as shown in Table~\ref{feature_Analysis}, the top three features that consistently appear in the solutions are 21, 20, and 27, corresponding to ``Concave Point Worst'', ``Fractal Dimension'', and ``Radius Worst'' respectively. The concave point worst feature indicates the severity of the concave portion of the shape, with ``worst'' denoting the highest mean value. The ``fractal dimension'' is a crucial characteristic that provides information related to the geometric shape of the fractals. The third feature, radius worst, represents the largest mean value for the distances from the centre to points on the perimeter.

While other ML models may share the feature of interpretability, they often present challenges that GE does not encounter. Decision trees and RF, though interpretable, lose clarity with complex structures and aggregation~\cite{arrieta2020explainable}. LDA relies on the Gaussian distribution of the data and assumes that the covariance of two classes is the same~\cite{ghojogh2019linear}, limiting its applicability. In contrast, GE does not depend on these factors and maintains transparency throughout its evolution, even when addressing complex and non-linear problems.

\begin{table}[]
\setlength{\tabcolsep}{6pt}
\caption{This analysis unveils prevalent features used by GE in all five setups. For $S_{CC}$ and $S_{MLO}$, percentages are computed from 8684 and 7945 occurrences. Likewise, contributions to $S_{CC + MLO}$ and $F_{CC + MLO}$ are based on 8138 and 8522 occurrences, respectively. The features of \textit{WBC} are also examined, with percentages drawn from 9076 appearances.}
\label{feature_Analysis}
\resizebox{\columnwidth}{!}{%
\begin{tabular}{cc@{\hspace{24pt}}cc@{\hspace{24pt}}cc@{\hspace{24pt}}cc@{\hspace{12pt}}@{\hspace{12pt}}cc}
\hline
\multicolumn{2}{c}{$S_{CC}$} & \multicolumn{2}{c}{$S_{MLO}$} & \multicolumn{2}{c}{$S_{CC+MLO}$} & \multicolumn{2}{c}{$F_{CC+MLO}$} & \multicolumn{2}{c}{\textit{WBC}} \\ \hline
Feature       & Usage        & Feature        & Usage        & Feature         & Usage         & Feature         & Usage          & Feature         & Usage          \\
17            & 6.22\%       & 5              & 4.93\%       & 17              & 5.35\%        & 41              & 3.78\%         & 21              & 7.83\%         \\
41            & 4.87\%       & 4              & 4.46\%       & 5               & 4.36\%        & 37              & 3.71\%         & 20              & 7.64\%         \\
38            & 4.58\%       & 41             & 3.95\%       & 7               & 4.33\%        & 4               & 3.63\%         & 27              & 6.47\%         \\
5             & 4.19\%       & 7              & 3.75\%       & 41              & 4.02\%        & 38              & 3.46\%         & 24              & 5.56\%         \\
18            & 3.88\%       & 17             & 3.65\%       & 18              & 3.93\%        & 11              & 3.38\%         & 1               & 5.1\%          \\
7             & 3.50\%       & 34             & 3.34\%       & 38              & 3.55\%        & 17              & 3.18\%         & 13              & 4.87\%         \\
36            & 2.73\%       & 45             & 3.15\%       & 11              & 3.08\%        & 5               & 3.11\%         & 7               & 4.23\%               \\ \hline
\end{tabular}%
}
\end{table}

\subsection{Statistical Analysis}

The statistical comparison of implemented data augmentation techniques involved a non-parametric Bayesian signed-rank test~\cite{benavoli2014bayesian} applied to each dataset. In our analysis, conducted on nine augmentation techniques with 30 paired AUC samples each, the test distinguished between methods being pair-wise larger, smaller or inconclusive. The approaches listed in the rows are compared with the methods presented in the corresponding column. The subsequent Bayesian signed-rank test revealed significant distinctions among the techniques. In the cases where STEM has outperformed the other approaches are underlined in the Table~\ref{bayseian test}. 

In the $S_{CC}$ setup, as illustrated in Table~\ref{bayseian test}(a), STEM, Mixup, SMOTE, ADA, S-NC, SVM-S, S-Tomek and BSMOTE all exhibit larger medians than S-ENN.

The statistical comparison of medians depicted in Table~\ref{bayseian test}(b) among various augmentation populations reveals notable differences for $S_{MLO}$ setup. STEM, S-NC, S-Tomek, ADA, and Mixup exhibit larger medians compared to BSMOTE, SVM-S, SMOTE, and S-ENN.

Similarly, for setup $S_{CC+MLO}$ in the Table~\ref{bayseian test}(c) STEM again showcases its effectiveness by outperforming S-NC, BSMOTE, Mixup, SMOTE, and ADA in medians. Additionally, S-ENN demonstrates superiority by exhibiting larger medians than Mixup, SMOTE, and ADA. Additionally, S-Tomek outperforms SMOTE in median values.  SVM-S, in particular, stands out with a larger median than ADA.

Moreover, STEM stands out by consistently surpassing S-Tomek, Mixup, BSMOTE, ADA, SVM-S, SMOTE, and S-ENN in median values presented in Table~\ref{bayseian test}(d) for $F_{CC+MLO}$ . Additionally, S-NC demonstrates superiority over SMOTE and S-ENN, while S-Tomek outperforms S-ENN in median values. Mixup, BSMOTE, ADA, SVM-S, and SMOTE all exhibit larger medians than S-ENN.

Finally, in the \textit{WBC} setup, as depicted in Table~\ref{bayseian test}(e), STEM emerged as the top-performing method, surpassing S-NC, BSMOTE, S-Tomek, Mixup, SVM-S, SMOTE, and ADA. S-NC exhibited a higher median than SMOTE and ADA, while Mixup outperformed SMOTE in median value. SVM-S demonstrated a larger median than SMOTE and ADA.

\begin{table}[]
\caption{The results of the Bayesian signed-ranked test are summarized here for the nine augmentation approaches for each data setup. Arrows indicate the direction of differences: $\Uparrow$ for larger, $\Downarrow$ for smaller, - for inconclusive, and N/A for not applicable results. A family-wise significance level of $\alpha\equiv 0.05$ is employed.}
\label{bayseian test}
\resizebox{\columnwidth}{!}{%
\begin{tabular}{cccccccccc}
\hline
\multicolumn{10}{c}{(a) $S_{CC}$}                                                                                                              \\ \hline
        & STEM       & Mixup      & SMOTE        & ADA          & S-NC         & SVM-S        & S-Tomek      & BSMOTE       & S-ENN        \\ \hline
STEM    & N/A        & -          & -            & -           & -            & -            & -            & -            & \boldmath$\underline\Uparrow$\\
Mixup   & -          & N/A        & -            & -            & -            & -            & -            & -            & $\Uparrow$ \\
SMOTE   & -          & -          & N/A          & -            & -            & -            & -            & -            & $\Uparrow$ \\
ADA     & -          & -          & -            & N/A          & -            & -            & -            & -            & $\Uparrow$ \\
S-NC    & -          & -          & -            & -            & N/A          & -            & -            & -            & $\Uparrow$ \\
SVM-S   & -          & -          & -            & -            & -            & N/A          & -            & -            & $\Uparrow$ \\
S-Tomek & -          & -          & -            & -            & -            & -            & N/A          & -            & $\Uparrow$ \\
BSMOTE  & -          & -          &              & -            & -            & -            & -            & N/A          & $\Uparrow$ \\
S-ENN   & $\Downarrow$ & $\Downarrow$ & $\Downarrow$   & $\Downarrow$  & $\Downarrow$   & $\Downarrow$   & $\Downarrow$  & $\Downarrow$  & N/A          \\ \hline
\multicolumn{10}{c}{(b) $S_{MLO}$}                                                                                                             \\ \hline
        & STEM       & Mixup      & S-NC         & S-Tomek      & ADA          & BSMOTE       & SVM-S        & SMOTE        & S-ENN        \\ \hline
STEM    & N/A        & -          & -            & -            & -            & \boldmath$\underline\Uparrow$ &\boldmath$\underline\Uparrow$ & \boldmath$\underline\Uparrow$ & \boldmath$\underline\Uparrow$ \\
Mixup   & -          & N/A        & -            & -            & -            & $\Uparrow$ & $\Uparrow$ & $\Uparrow$ & $\Uparrow$ \\
S-NC    & -          & -          & N/A          & -            & -            & $\Uparrow$ & $\Uparrow$ & $\Uparrow$ &$\Uparrow$ \\
S-Tomek & -          & -          & -            & N/A          & -            & $\Uparrow$ & $\Uparrow$ & $\Uparrow$ & $\Uparrow$ \\
ADA     & -          & -          & -            & -            & N/A          & $\Uparrow$ & $\Uparrow$ & $\Uparrow$ & $\Uparrow$ \\
BSMOTE  & $\Downarrow$& $\Downarrow$& $\Downarrow$  & $\Downarrow$   & $\Downarrow$  & N/A          & -            & -            & -            \\
SVM-S   & $\Downarrow$ & $\Downarrow$ & $\Downarrow$   & $\Downarrow$   & $\Downarrow$   & -            & N/A          & -            & -            \\
SMOTE   & $\Downarrow$ & $\Downarrow$ & $\Downarrow$  &$\Downarrow$  & $\Downarrow$   & -            & -            & N/A          & -            \\
S-ENN   & $\Downarrow$ & $\Downarrow$ & $\Downarrow$   & $\Downarrow$  &$\Downarrow$   & -            & -            & -            & N/A          \\ \hline
\multicolumn{10}{c}{(c) $S_{CC+MLO}$}                                                                                                          \\ \hline
        & STEM       & S-ENN      & S-Tomek      & SVM-S        & S-NC         & BSMOTE       & Mixup        & SMOTE        & ADA          \\ \hline
STEM    & N/A        & -          & -            & -            & -            & \boldmath$\underline\Uparrow$& \boldmath$\underline\Uparrow$ & \boldmath$\underline\Uparrow$ & \boldmath$\underline\Uparrow$ \\
S-ENN   & -          & N/A        & -            & -            & -            & -            &  $\Uparrow$ &  $\Uparrow$ &  $\Uparrow$ \\
S-Tomek & -          & -          & N/A          & -            & -            & -            & -            & -          & -            \\
SVM-S   & -          & -          & -            & N/A          & -            & -            & -            & -            &  $\Uparrow$\\
S-NC    & -          & -          & -            & -            & N/A          & -            & -            & -            & -            \\
BSMOTE  &  $\Downarrow$  & -          & -            & -            & -            & N/A          & -            & -            & -            \\
Mixup   &  $\Downarrow$  & $\Downarrow$  & -            & -            & -            & -            & N/A          & -            & -            \\
SMOTE   &  $\Downarrow$  &  $\Downarrow$  & -            & -            & -            & -            & -            & N/A          & -            \\
ADA     &  $\Downarrow$  &  $\Downarrow$  & -            &  $\Downarrow$    & -            & -            & -            & -            & N/A          \\ \hline
\multicolumn{10}{c}{(d) $F_{CC+MLO}$}                                                                                                          \\ \hline
        & STEM       & S-NC       & Mixup        & S-Tomek      & BSMOTE       & ADA          & SVM-S        & SMOTE        & S-ENN        \\ \hline
STEM    & N/A        & -          & \boldmath$\underline\Uparrow$& \boldmath$\underline\Uparrow$& \boldmath$\underline\Uparrow$ & \boldmath$\underline\Uparrow$ & \boldmath$\underline\Uparrow$ & \boldmath$\underline\Uparrow$ & \boldmath$\underline\Uparrow$ \\
S-NC    & -          & N/A        & -            & -            & -            & -            & -            & -            & $\Uparrow$ \\
Mixup   & $\Downarrow$  & -          & N/A          & -            & -            & -            & -            & -            & $\Uparrow$ \\
S-Tomek &$\Downarrow$  & -          & -            & N/A          & -            & -            & -            & -            & $\Uparrow$ \\
BSMOTE  & $\Downarrow$  & -          & -            & -            & N/A          & -            & -            & -            & $\Uparrow$\\
ADA     & $\Downarrow$  & -          & -            & -            & -            & N/A          & -            & -            & $\Uparrow$ \\
SVM-S   & $\Downarrow$  & -          & -            & -            & -            & -            & N/A          & -            &$\Uparrow$ \\
SMOTE   & $\Downarrow$  & -          & -            & -            & -            & -            & -            & N/A          & $\Uparrow$ \\
S-ENN   & $\Downarrow$  &$\Downarrow$  & $\Downarrow$   & $\Downarrow$    & $\Downarrow$   & $\Downarrow$   &$\Downarrow$   & $\Downarrow$   & N/A          \\ \hline
\multicolumn{10}{c}{(e) \textit{WBC}}                                                                                                          \\ \hline
        & STEM       & S-ENN      & S-NC         & SVM-S        & Mixup        & ADA          & BSMOTE       & S-Tomek      & SMOTE        \\ \hline
STEM    & N/A        & -          & \boldmath$\underline\Uparrow$ & \boldmath$\underline\Uparrow$ & \boldmath$\underline\Uparrow$ & \boldmath$\underline\Uparrow$ & \boldmath$\underline\Uparrow$ & \boldmath$\underline\Uparrow$ & \boldmath$\underline\Uparrow$ \\
S-ENN   & -          & N/A        & $\Uparrow$  & $\Uparrow$  & $\Uparrow$  & $\Uparrow$  & -            & $\Uparrow$  & $\Uparrow$  \\
S-NC    & $\Downarrow$  & $\Downarrow$  & N/A          & -            & -            & -            & -            & -            & -            \\
SVM-S   & $\Downarrow$  & $\Downarrow$  & -            & N/A          & -            & $\Downarrow$  & -            & -            & $\Downarrow$  \\
Mixup   & $\Downarrow$  & $\Downarrow$  & -            & -            & N/A          & -            & -            & -            &$\Uparrow$ \\
ADA     & $\Downarrow$  & $\Downarrow$  & -            & $\Downarrow$    & -            & N/A          & -            & -            & -            \\
BSMOTE  & $\Downarrow$  & - & -            & -            & -            & -            & N/A          & -            & -            \\
S-Tomek & $\Downarrow$  & $\Downarrow$  & -            & -            & -            & -            & -            & N/A          & -            \\
SMOTE   & $\Downarrow$  & $\Downarrow$  & -            & $\Downarrow$    & $\Downarrow$    & -            & -            & -            & N/A          \\ \hline
\end{tabular}%
}
\end{table}

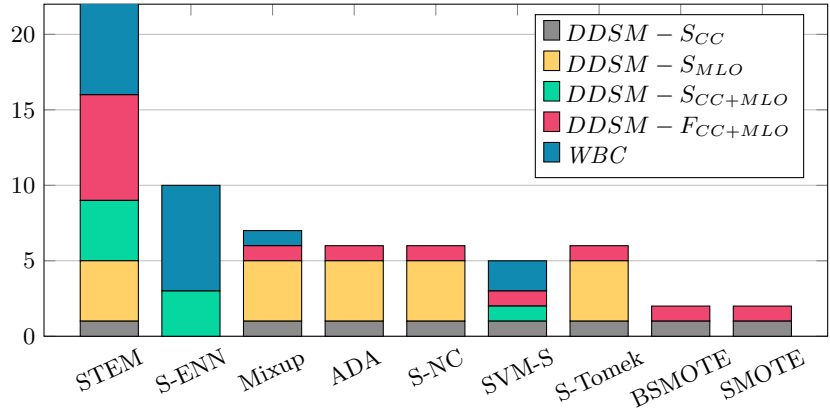
\begin{figure}[!ht]
\centering
\begin{tikzpicture}
\begin{axis}[
	ybar stacked,
	ymin=0,
	ymax=22,
	ymajorgrids = true,
	bar width=22pt,
	xtick=data,
	legend style={
    	cells={anchor=west},
    	legend pos=north east,
	},
	xticklabels from table={\testdata}{Label},
	xticklabel style={rotate=25,align=center},
]
	\addplot [fill=fuel,point meta=y]
    	table [y=s1, meta=Label, x expr=\coordindex]
        	{\testdata};
            	\addlegendentry{$DDSM - S_{CC}$}
	\addplot [fill=banana,point meta=y]
    	table [y=s2, meta=Label, x expr=\coordindex]
        	{\testdata};
            	\addlegendentry{$DDSM - S_{MLO}$}
	\addplot [fill=neptune,point meta=y]
    	table [y=s3, meta=Label, x expr=\coordindex]
        	{\testdata};
            	\addlegendentry{$DDSM - S_{CC+MLO}$}
	\addplot [fill=quincy]
    	table [y=s4, meta=Label, x expr=\coordindex]
        	{\testdata};
            	\addlegendentry{$DDSM - F_{CC+MLO}$}
	\addplot [fill=tango]
    	table [y=s5, meta=Label, x expr=\coordindex]
        	{\testdata};
            	\addlegendentry{\textit{WBC}}
\end{axis}
\end{tikzpicture}
\caption{The illustration of the overall results acquired from the Bayesian signed-rank test is shown here. The cumulative score is the total number of times one approach outperforms the other. STEM obtained a cumulative score of 23 where the maximum possible is 40 (comparing one versus another 8 approaches in 5 setups), outperforming the other approaches. Each color represents distinct test setups used for the evaluation.}
\label{rank_count}
\end{figure}

The Bayesian analysis results are summarized in Fig.~\ref{rank_count}. It reveals that STEM, a combination of S-ENN and Mixup, emerges as the top-ranking approach. This result underscores the effectiveness of this combined strategy in enhancing performance. Notably, S-ENN and Mixup individually secure the second and third positions, further affirming the significance of this ensemble approach.

\section{Conclusion and Future Work}
\label{Conclusion}
In this study, we addressed class imbalance and interpretability challenges in medical imaging diagnosis by using GE to produce classifier trained on data augmented by the recently-introduced STEM technique. Our approach not only delivers interpretable solutions but also outperforms an ensemble of other ML classifiers in terms of performance. The analysis conducted on the $DDSM$ and \textit{WBC} datasets emphasizes the effectiveness of GE, as evidenced by improvements in AUC and its ability to identify critical data features. Notably, our inclusion of Bayesian signed-rank test results confirms that STEM emerges as the best-performing approach for augmentation. The improved AUC and enhanced interpretability of our approach can help build trust and facilitate informed decisions. Thus, our study validates the proposed hypothesis, demonstrating the efficacy of the combined GE and STEM approach.

For future research, we suggest improving performance by incorporating additional image attributes, such as wavelet transformations and local binary patterns, to enhance the feature set and dataset diversity.  Furthermore, exploring the mixture of different datasets to assess the robustness of our approach across various image data sources would be interesting.

\section{ACKNOWLEDGEMENTS}
The Science Foundation Ireland (SFI) Centre for Research Training in Artiﬁcial Intelligence (CRT-AI), Grant No. 18/CRT/6223 and the Irish Software Engineering Research Centre (Lero), Grant No. 16/IA/4605, both provided funding for this study.


 \bibliographystyle{splncs04}
 \bibliography{bibliography}

\end{document}